# *Features of word similarity*

Arthur M. Jacobs[1,2] & Annette Kinder[3]

Author Note
1) Experimental and Neurocognitive Psychology, Freie Universität Berlin, Germany
2) Center for Cognitive Neuroscience Berlin (CCNB), Freie Universität Berlin, Germany
3) Learning Psychology, Freie Universität Berlin, Germany

Correspondence: Arthur M. Jacobs
Experimental and Neurocognitive Psychology, Freie Universität Berlin, Habelschwerdter Allee 45 , D-14195 Berlin, Germany.
Email: ajacobs@zedat.fu-berlin.de

**Abstract**

In this theoretical note we compare different types of computational models of word similarity and association in their ability to predict a set of about 900 rating data. Using regression and predictive modeling tools (neural net, decision tree) the performance of a total of 28 models using different combinations of both surface and semantic word features is evaluated. The results present evidence for the hypothesis that word similarity ratings are based on more than only semantic relatedness. The limited cross-validated performance of the models asks for the development of psychological process models of the word similarity rating task.

## Introduction

The general and central construct of similarity has a long tradition in psychology and other sciences and has generated both a multitude of relevant empirical data and intensive theoretical debate (e.g., Krumhansl, 1978; Medin et al., 1993; Rubenstein & Goodenough, 1965; Shepard, 1987; Tversky, 1977). The special case of word similarity seems to have attracted relatively little psychological research when compared with its prominent role in computational linguistics, natural language processing and machine learning (e.g., Baroni et al., 2014; Budanitzky & Hirst, 2006; Hill et al., 2014; Le & Fokkens, 2016; Turney et al., 2010; see Mandera et al., 2017, for review on psychological research). This is evidenced by the fact that the most recent and extensive empirical data bases concerning word similarity ratings seem to originate in the latter sciences, in particular the new 'gold-standard' called SimLex-999 (Hill et al., 2014). The theoretical debate in these fields centers around the notions of semantic similarity (e.g., 'car, bicycle') vs. association (e.g., 'car', 'factory'; cf. Hofmann et al., 2018; Turney, 2006), often inspired by psychological theories (e.g., Medin et al., 1990; Tversky, 1977), as well as embodied/experiential or knowledge-based semantics (e.g., Andrews et al., 2009; Hofmann & Jacobs, 2014; Jacobs et al., 2016) vs. distributional semantics (e.g., Le & Fokkens, 2016).

In this note, we attempt to advance interdisciplinary efforts aiming at integrating lexical and computational semantics (e.g., Hofmann et al., 2018; Hollis et al., 2017; Jurgens et al., 2012; Mandera et al., 2017) by testing different computational models against data from SimLex-999 (*SL999* henceforth). Extending previous research (Hill et al., 2015; Mandera et al., 2017) we develop and test computional models that not only include different similarity and association metrics but also a set of both surface and novel lexico-semantic features that have been used succesfully to predict the beauty of words (Jacobs, 2017) or the aptness and literariness of metaphors (Jacobs & Kinder, 2017, 2018). In particular, we investigate the hypothesis that word similarity ratings are based on more features than only semantic relatedness.

### Computational Models of Word Similarity

Two types of models for predicting human word similarity ratings dominate the literature: knowledge-based, dictionary (DICT) models (henceforth called type I) and corpus-based, vector space models (VSM, henceforth called type II). Type I models are based on a taxonomy using hierarchies of hypernym-hyponym ('is a') relations (i.e., wordnet; Fellbaum, 1988) and can be subdivided into 'pure' taxonomic models, type I A (e.g., Leacock & Chodorow, 1998; Wu & Palmer, 1994) and hybrid ones (type I B) that also make use of corpus-based information (Jiang & Conrath, 1997; Lin, 1998; Resnik, 1995). Type II models are based on distributional semantics and can be subdivided into count models (type II A; e.g., latent semantic analysis/LSA; Deerwester et al., 1990) and predict models (type II B; word2vec/w2v; Baroni et al., 2014; cf. Mandera et al., 2017). Both subtypes of models rely on word-context relations: Count models use information on the number of occurrences of words in context whereas predict models are neural nets trained on predicting words from their context or vice versa. To what extent such models are psychologically more plausible has recently been discussed by Mandera et al. (2017) who opted in favor of predict models.

**Present Working Model**

The above type I and II models all (implicitly) assume that word similarity ratings are based on semantic similarity or association only. However, how plausible is it that readers exclusively use this kind of semantic information –be it experiential, distributed or some mixture of the two– when rating word similarity? After all, the literature presents ample evidence for the fact that orthographic, phonological and other semantic features (e.g., affective semantics; e.g., Jacobs et al., 2015, 2016; Warriner et al., 2013) play a role in (single) word recognition (e.g., Andrews et al., 2009; Jacobs & Grainger, 1994). Thus, to the extent that the similarity rating task involves the automatic recognition of two words –preceding the rating– it seems plausible that these factors also influence the ratings. Our working model thus assumes that a similarity rating is based on a set of variables representing both surface (i.e., orthographic-phonological) and semantic features of the word pair. Each similarity variable, in turn, is a compound of N features. For example, orthographic similarity is often computed using the orthographic neighborhood density and frequency, two variables known to influence word recognition performance (e.g., Grainger & Jacobs, 1996). Phonological similarity can be computed based on the number and nature of syllables or the consonant-vowel quotient. In addition to the semantic measures provided by type I and II models, semantic similarity can be based on a multitude of features including affective semantics (e.g., valence, arousal, emotion potential, Bestgen & Vincze, 2012; Jacobs, 2015; Jacobs et al., 2015; Westbury et al., 2015), embodied semantics (e.g., Juhasz et al., 2011; Siakaluk et al., 2008), feature norms (e.g., McRae et al., 2005), or aesthetic semantics (e.g., sonority score, aesthetic potential; Jacobs, 2017; Jacobs & Kinder, 2018).

In line with this working model, we introduce a third type of computational model for word similarity ratings based on a set of surface and lexico-semantic features. Since these features are extracted via methods used in Quantitative Narrative Analysis (QNA; e.g., Graesser et al., 2004; Jacobs et al., 2016; Jacobs & Kinder, 2017; Pennebaker & Francis, 1999), we coin them QNA models (type III). Type III models combine a multitude of relevant (sub)lexical psycholinguistic features known to influence word recognition and reading from different sources. Finally, a fourth class, hybrid models (type IV), are considered here which combine features from different model types, e.g. type II and III.

As an 'exploratory test' of the hypothesis that word similarity ratings are based on several sets of features we computed regressions between the ouput of the type I and II models as compared to that of several QNA (type III) and increasingly complex hybrid (type IV) models[1]. The correlational data typically used in benchmark studies (Le & Fokkens, 2016; Bojanowski et al., 2017, Hill et al., 2015) provide no real predictions (from a training to a test set), i.e. they usually are not cross-validated (cf. Faruqui et al., 2016). Therefore, we also carried out cross-validated *predictive modeling* analyses for all models (cf. Jacobs 2017; Jacobs et al., 2016, 2017; Jacobs & Kinder, 2017, 2018; Mandera et al., 2015).

**Methods**

Similar to Le and Fokkens (2016), we calculated the similarity scores of all noun and verb pairs in SimLex-999 (a set of 888 pairs downloaded from: https://www.cl.cam.ac.uk/~fh295/simlex.html) using the models described in Table 1. The type I models were computed via the NLTK.wordnet python package (Bird et al., 2009). The wordnet database provides information about ~120.000 words/concepts. For the type II models, both the LSA and w2v measures were computed using the gensim python package (Rehurek & Sojka, 2010). For comparison we used a small vs. a large training corpus (Brown vs. wiki-news) and varied the number of dimensions. The LSABrown10 and LSABrown100 models were computed with either 10 or 100 dimensional vectors using the Brown corpus (Kucera & Francis, 1965). The w2vb200 model was computed from the same corpus with 200 dimensions. For the w2vwiki300 model we used the 300d vectors enriched with subword information available at: https://fasttext.cc/docs/en/english-vectors.html (Bojanowski et al., 2016). To ensure that our conclusions do not depend on the choice of a specific regressor model, we used two nonlinear tools for our predictive modeling analyses: a neural net (i.e., MultiLayerPerceptron/MLP; Rumelhart et al., 1986) and a potent decision tree regressor (i.e., ExtraTreesRegressor/ERT; Geurts et al., 2006), as well as multiple linear regression/MLR. The QNA features of the type III and IV models were computed based on python scripts used in previous QNA studies (Jacobs, 2017; Jacobs & Kinder, 2017, 2018), using the differences between these measures for w1 and w2 as a (*dis-)*similarity measure. All variables were standardized before the computations. The models were computed using python scripts based on the sklearn package (Pedregosa et al., 2011) using the ERT, MLP or MLR regressors (http://scikit-learn.org/stable/supervised_learning.html; for discussion of the advantages of classifiers/regressors of the decision tree family, see Strobl et al., 2009). The parameters for the regressors were tuned to the entire data set via grid-search and thus could vary for each model[2]. The training and test sets for cross-validation consisted of 666 and 222 randomly drawn items, respectively. To maximise data reliability we ran 1000 iterations for each regressor and model and averaged the results.

[1] Given the complexity and practical non-comparability of the models considered here, we renounced on inference statitistical model evaluation. For example, the number of parameters that can, in principle, be varied in the corpus-based models discourages any formal comparisons as much as other aspects that would have to be taken into account such as training corpus size (and quality) which cannot be meaningfully compared between type I and II models (cf. Jacobs, 2018a,b). Similarly, the present analyses represent no contest between different regressor models (i.e., ERT, MLP, MLR): the fact that, descriptively, the ERT regressor overall appears to achieve the best fits does not mean that this would also hold for a different data set.

**Results**

The results of our comparative exploratory computational study using 28 different models implementing a wide range of word features (total = 23) to predict the SL999 word similarity ratings are summarized in Table 1 which also briefly describes the model specifics and indicates model performance with the $R^2$ index.

[2] A typical parameter set for the ERT models was: number of trees in forest = 50, minimum samples per split = 3, maximum number of features = auto, maximum depth of tree = None, minimum samples leaf = 1, bootstrap = False; and for the MLP models: number of neurons per hidden layer (1 or 2) = 10 + 10, activation function = hyperbolic tan (tanh), solver = stochastic gradient descent, learning rate = adaptive, initial learning rate = 0.001, early stopping = True. The MLP models were occasionally boosted using AdaBoost when there was no convergence after 500 runs.

**Table 1. Models with specifics, computation examples, performance index and most important features**

| Model # (metrics, authors, corpus) | word pair similarity example, z-value | brief description | $R^2_{corr}$ | Most important features |
|---|---|---|---|---|
| Type Ia. DICT-pure | | | | |
| 1. *wn-path length* | *car-bicycle: .1*<br>*car-factory: -.89* | path distance, i.e. shortest number of edges (d) between wordnet synsets, $s_1$, $s_2$ in 'is a' relation hierarchy<br>wn-path = $1/[d(s_1,s_2)+1]$,<br>e.g. d(car,bicycle) = 2; d(car,factory) = 7 | .25/.27 | - |
| 2. *wn-lch*<br>Leacock & Chodorow (1998) | *car-bicycle: .61*<br>*car-factory: -1.1* | normalizes path-based scores by the maximum depth (D) of the hierarchy, e.g. D(car) = 9<br><br>wn-lch = $-\log[(d(s_1,s_2)+1)/2D]$ | .29/.32 | - |
| 3. *wn-wup*<br><br>Wu & Palmer (1994) | *car-bicycle: .56*<br>*car factory: -.2* | takes into account that senses deeper in the hierarchy tend to be more specific than those high up. L = least common subsumer, i.e. most specific concept shared by two words [L(car,bicycle) = 'wheeled-vehicle'; L(car,factory) = 'structure'].<br><br>wn-wup = $2D[L(s_1,s_2)]/[D(s_1)+D(s_1)]$ | .21/.21 | - |
| Type Ib. DICT-hybrid | | | | |
| 4. *wn-res*<br><br>Resnik (1995) | *car-bicycle: .7*<br>*car-factory: -.13* | takes information content (IC) into account;<br>IC(s) = -log[f(s)], p(s) = frequency of $s_1$ or $s_2$ in Brown corpus; e.g. IC(car) = 8.54<br><br>wn-res = $IC[L(s_1, s_2)]$ | .21/.22 | - |
| 5. *wn-jcn*<br>Jiang & Conrath (1997) | *car-bicycle:* -.35<br>*car-factory:* 1.35 | also takes information content (IC) into account;<br>wn-jcn = $IC(s_1) + IC(s_1) - 2IC[L(s_1, s_2)]$ | .006/.0 05 | - |
| 6. *wn-lin*<br>Lin (1998) | *car bicycle: .57*<br>*car-factory: -1.16* | also takes information content (IC) into account;<br>IC = 'ic-semcor.dat' corpus, i.e. frequency of occurrence of $s_1$ or $s_2$ in collection of sense-tagged text used by wordnet<br><br>2wn-res / $[IC(s_1) + IC(s_1)]$ | .25/.28 | - |
| **7.** ***Type Iall*, N = 6** | ***car bicycle: .36***<br>***car-factory: -1.1***<br>***ERT*** | **all six metrics together** | **.71/.31**<br>**ERT**<br>**.36/.31** | **lin,path,lch** |
| | ***car bicycle: .32***<br>***car-factory: -.93***<br>***MLP***<br>***car bicycle: .54***<br>***car-factory: -1.1***<br>***MLR*** | | **MLP**<br>**.35/.33**<br>**MLR** | |
| Type IIa. VSM-count | | | | |
| 8. *LSAb10*<br><br>9. *LSAb100*<br>*Count* model: *LSA* (Deerwester et al., 1990)<br>Training corpus = Brown (~1Mio words; Kucera & Francis, 1965) | *car-bicycle: .89*<br>*car-factory: .41*<br><br>*car-bicycle: -1.16*<br>*car-factory: -.4* | counts how many times a word appears in a document resulting in a document-term matrix (DTM) which is then decomposed into the product of three other matrices using dimensionality reduction (singular value decomposition).<br><br>Each word is represented by an n-dim vector of factor weights extracted from the DTM<br><br>semantic similarity/association measure = cosine of word vectors | .0001/-.004 (LSAb10)<br><br>.002/-.013 (LSAb100, n.s.) | - |
| Type IIb. VSM-predict | | | | |
| 10. *w2vb200*<br><br>11. *w2vwiki300*<br>*Predict* model: (Mikolov et al., 2013) CBOW and Skip-Gram/SG<br>Training corpus = wiki-news-300d-1M-subword.vec (~1Mio word types; Bojanowski et al., 2016) | *car-bicycle: -.7*<br>*car-factory: -1.2*<br><br>*car-bicycle: .35*<br>*car-factory: -.3* | simple (1 hidden layer) neural network model learns to predict the current word (t) given context words (c, Continuous Bag of Words model, CBOW) or c given t (skip-gram model, SG) based on a relatively narrow window (usually ±5 words) that slides through a training corpus. Thus, e.g. SG captures the distribution p(c\|t) in the window computing normalized probabilities via negative sampling (i.e., by having each training sample only modify a small percentage of the weights). For each pair (c,t) taken from training data, c is replaced by random words drawn from vocabulary to obtain new pairs {(c′, t)}.<br><br>SG identifies which pairs come from a positive distribution (M) and which from a negative (N), maximizing the negative log likelihood:<br>l = -∑log p(M\|c,t) + -∑log p(N\|c',t) | .013/.022 (*w2vb200*)<br><br>.22/.22 (*w2vwiki300*) | - |
| **12.** ***Type IIall*, N = 4 (but: 10d + 100d + 200d + 300d vectors)**[3] | ***car bicycle: .06***<br>***car-factory: -.53***<br>***ERT***<br>***car bicycle: .71***<br>***car-factory: -.24***<br>***MLP***<br>***car bicycle: .6***<br>***car-factory: -.21***<br>***MLR*** | ***LSAb10 + LSAb100 + w2vb200 + wiki300sub*** | **.89/.32**<br>**ERT**<br>**.27/.30**<br>**MLP**<br>**.25/.24**<br>**MLR** | **w2vwiki300, w2vb200 LSAb10, LSAb100** |
| Type III[4]. QNA | | | | |
| 13. *Surface/Surf*, N = 5 | *car-bicycle: -.73*<br>*car-factory: -.8*<br>*ERT* | Five surface features used in the literature on word recognition: *number of letters/nlet, log frequency Zipf/logfZ* (van | .12/.03<br>ERT | cvq,on,nlet,nsyl logfZ |

3 It is of note that a direct comparison of the number of features (N) can be misleading. While in the surface model each word is decribed by a vector of five QNA features, it is only one composite feature e.g. in the *w2vb200* model; but this ‚single' feature is based on a 200d vector.

4 The results for the single surface/AS features were: nlet: .005/-.009, nsyl: .005/-.005, on: .007/-.02, logfZ: .006/-.005 cvq: .001/.001, val: .0006/-.005, aro: .0001/-.006, ima: .0007/-.0008, dom: .0004/-.005, conc: .003/.003, dist: .001/-.02 (all n.s).

| | | | | |
|---|---|---|---|---|
| | *car-bicycle:* -1.9<br>*car-factory: -1.8*<br>*MLP*<br>*car-bicycle:* -1.4<br>*car-factory: -1.3*<br>*MLR* | Heuven et al., 2014), *consonant-vowel quotient/cvq* (Jacobs, 2017)*, orthographic neighborhood density/on* (Coltheart, 1981), *number of syllables/nsyl*<br><br>input to the models are the differences between these measures for w1 and w2 | .04/.03<br>MLP<br><br>.008/<br>-.01<br>MLR | |
| 14. *Affective-semantic/AS*, N = 6 | *car-bicycle:* -.43<br>*car-factory: -.77*<br>*ERT*<br>*car-bicycle:* -.3<br>*car-factory: -.71*<br>*MLP*<br>*car-bicycle:* -.34<br>*car-factory: -.14*<br>*MLR* | five features from the Bestgen and Vincze (2012) database: *valence/val, arousa/arol, imageability/ima, dominance/dom, concreteness/conc* plus one novel feature: *distinctiveness/dist* (the inverse of the mean semantic similarity of each target word with each word in the wiki300sub corpus was computed and then the difference was calculated for each of the 888 pairs).<br><br>input/output as in model 13 | .43/.1<br>ERT<br><br>.18/.14<br>MLP<br><br>.009/<br>-.02<br>MLR | aro,conc,val, ima,dom,dist |
| 15. *Aesthetic/AEST*, N = 4 | *car-bicycle:* -.4<br>*car-factory: -.77*<br>*ERT*<br>*car-bicycle:* .12<br>*car-factory: -.44*<br>*MLP*<br>*car-bicycle:* -1.0<br>*car-factory: -.57*<br>*MLR* | previous studies suggest that four features play a role in aesthetic ratings of single words (e.g., Jacobs, 2017; Jacobs et al., 2015; 2016): *logfZ* (as as index of familarity)*, valence, sonority score/sc, aesthetic potential/ap.* | .65/.01<br>ERT<br>.03/.05<br>MLP<br>.0008/<br>-.009<br>MLR | ap,logfZ,val,sc |
| 16. *Surf* + *AS*, N = 11 | *car-bicycle:* -.72<br>*car-factory: -.87*<br>*ERT*<br>*car-bicycle:* -1.9<br>*car-factory: -1.8*<br>*MLP*<br>*car-bicycle:* -1.2<br>*car-factory: -1.1*<br>*MLR* | models 13 and 14 combined | .82/.1<br>ERT<br>.03/.01<br>MLP<br>.018/<br>-.019<br>MLR | conc,ima,aro, val,logfZ |
| 17. Surf + AEST, N = 7 | *car-bicycle:* -.6<br>*car-factory: -.8*<br>*ERT*<br>*car-bicycle:* -1.3<br>*car-factory: -1.0*<br>*MLP*<br>*car-bicycle:* -1.3<br>*car-factory: -1.5*<br>*MLR* | models 13 and 15 combined | .72/.02<br>ERT<br>.02/-.01<br>MLP<br>.01/<br>-.02<br>MLR | ap,logfZ,on,cvq, nsyl,sc,nlet |
| 18. AS + AEST, N = 8 | *car-bicycle:* -.5<br>*car-factory: -.66*<br>*ERT*<br>*car-bicycle:* .87<br>*car-factory: .33*<br>*MLP*<br>*car-bicycle:* -.37<br>*car-factory: -.12*<br>*MLR* | models 14 and 15 combined | .78/.02<br>ERT<br>.04/-<br>.005<br>MLP<br>.01/<br>-.02<br>MLR | conc,ap,aro,ima,dom val,dist |
| **19. *Surf* + *AS* + *AEST*, N = 13** | *car-bicycle:* -.7<br>*car-factory: -.9*<br>*ERT*<br>*car-bicycle:* -.73<br>*car-factory: -1.2*<br>*MLP*<br>*car-bicycle:* -1.2 | models 13, 14 and 15 combined | **.90/.11<br>ERT<br>.11/.00<br>2 MLP<br>.02/-.03<br>MLR** | **conc,ap** |
| | *car-factory: -1.1*<br>*MLR* | | | |
| Type IV. HYBRID | | | | |
| 20. Hybrid 1. *Type Iall + surf*, N = 11 | *car-bicycle:* .17<br>*car-factory: -.71*<br>*ERT*<br>*car-bicycle:* .13<br>*car-factory: -.85*<br>*MLP*<br>*car-bicycle:* .53<br>*car-factory: -1.2*<br>*MLR* | models 7 and 13 | .83/.33<br>ERT<br>.34/.31<br>MLP<br>.35/.33<br>MLR | lin,path,lch<br>logfZ |
| 21. Hybrid 2. *Type Iall + AS*, N = 12 | *car-bicycle:* -.06<br>*car-factory: -.78*<br>*ERT*<br>*car-bicycle:* .63<br>*car-factory: -1.17*<br>*MLP*<br>*car-bicycle:* .51<br>*car-factory: -1.2*<br>*MLR* | models 7 and 14 | .86/.38<br>ERT<br>.35/.33<br>MLP<br>.35/.33<br>MLR | lin,path,lch,dist |
| 22. Hybrid 3. *Type Iall + surf + AS*, N = 17 | *car-bicycle:* .01<br>*car-factory: -.74*<br>*ERT*<br>*car-bicycle:* .62<br>*car-factory: -1.2*<br>*MLP*<br>*car-bicycle:* .46<br>*car-factory: -1.2*<br>*MLR* | models 7, 13, and 14 combined | .90/.42<br>ERT<br>.36/.33<br>MLP<br>.36/.33<br>MLR | lin,path,lch |
| 23. Hybrid 4. *Type Iall + surf + AS + AEST*, N = 19 | *car-bicycle:* -.04<br>*car-factory: -.7*<br>*ERT*<br>*car-bicycle:* .29<br>*car-factory: -1.12*<br>*MLP*<br>*car-bicycle:* .49<br>*car-factory: -1.28*<br>*MLR* | models 7, 13, 14 and 15 combined | .9/.49<br>ERT<br>.28/.36<br>MLP<br>.36/.32<br>MLR | lin,path |
| 24. Hybrid 5. *Type IIall + surf*, N = 9 | *car-bicycle:* -.29<br>*car-factory: -.55*<br>*ERT*<br>*car-bicycle:* .71<br>*car-factory: -.43*<br>*MLP*<br>*car-bicycle:* .58<br>*car-factory: -.25*<br>*MLR* | models 12 and 13 | .82/.34<br>ERT<br>.26/.26<br>MLP<br>.26/.21<br>MLR | w2vwiki300,<br>logfZ |
| 25. Hybrid 6. *TypeIIall + AS*, N = 10 | *car-bicycle:* -.17<br>*car-factory: -.39*<br>*ERT*<br>*car-bicycle:* .71<br>*car-factory: -.37*<br>*MLP*<br>*car-bicycle:* .55<br>*car-factory: -.25*<br>*MLR* | models 12 and 14 | .92/.34<br>ERT<br>.25/.31<br>MLP<br>.26/.23<br>MLR | w2vwiki300 |
| 26. Hybrid 7. *TypeIIall + surf + AS*, N = 15 | *car-bicycle:* -.23<br>*car-factory: -.46*<br>*ERT*<br>*car-bicycle:* .49<br>*car-factory: -.26*<br>*MLP*<br>*car-bicycle:* .51<br>*car-factory: -.31* | models 12, 13, and 14 combined | .88/.31<br>ERT<br>.3/.24<br>MLP<br>.26/.22<br>MLR | w2vwiki300 |
| | *MLR* | | | |
| 27. Hybrid 8. *TypeIIall* + *surf* + *AS* + *AEST*, N = 17 | *car-bicycle:* -.24<br>*car-factory: -.6*<br>*ERT*<br>*car-bicycle:* .15<br>*car-factory: -.09*<br>*MLP*<br>*car-bicycle:* .51<br>*car-factory: -.3*<br>*MLR* | models 12, 13, 14 and 15 combined | .89/.31<br>ERT<br>.34/.3<br>MLP<br>.26/.22<br>MLR | w2vwiki300 |
| **28. Hybrid 13. *Type Iall* + *TypeIIall* + *surf* + *AS* + *AEST*, N = 23** | *car-bicycle:* -.04<br>*car-factory: -.65*<br>*ERT*<br>*car-bicycle:* .64<br>*car-factory: -.98*<br>*MLP*<br>*car-bicycle:* .58<br>*car-factory: -.95*<br>*MLR* | models 7, 12, 13, 14 and 15 combined | **.93/.47<br>ERT<br>.54/.41<br>MLP<br>.45/.35<br>MLR** | **lin,w2vwiki300<br>path,lch** |

**Notes**. Descriptively best performance in **bold**. $R^2_{corr}$: 1st value = for training set, 2nd value = test set. Feature ranks only for importance values >.1 according to best-fitting model.

**Figure 1 here**

## Discussion

The data in Table 1 present the following picture. First, regarding the six type I models only the *wn-jcn* model appears to stand out from the typical performance range established in the literature ($R^2_{corr}$ = .2 - .3[5]), with the *wn-lch* model being the winner, descriptively. The $R^2$s of the combined type I model (model 7) with lin, path and lch being the top-ranked metrics suggests that combining pure DICT models (e.g., Leacock & Chodorow, 1998) with hybrid DICT ones using corpus-based information (e.g., Lin, 1998) does not bring advantages.

Second, as concerns the corpus-based type II models (LSA and w2v), only the *w2vwiki300* model seems to play in the league of the type I models ($R^2_{corr/cv}$ = .22). Actually, the accuracy of this model is a bit higher than the w2v300d model used by Le and Fokkens (2015, $R^2_{corr}$ = .17, Table 1) who had already concluded that 'taxonomy-based approaches are more suitable to identify similarity'…'but that the inferior performance of corpus-based approaches may not (always) matter'. Although the superior $R^2$s of the combined type II model (model 12) suggests that when using, for example, a neural net approach for solving nonlinear multivariate problems corpus-based models may fare as well as type I models in accounting for similarity ratings, model 12 does not seem to be better than the combined type I model (model 7) or the wn-lch model (model 2). This can be seen as support for Le and Fokkens (2016) conclusion.

[5] The data for the wn-wup and wn-lch measures correspond almost exactly to those published by Le and Fokkens (2015, Table 1) who did not test the other four measures and did not use a training-test cross-validation.

As regards type III, none of the single-feature models reached a significant result. Of the combined models even model 19 with 13 features descriptively fared not as well as the best ‘single’-feature type I and II models, suggesting that –at least for the present feature set and empirical data– similarity ratings were not based on surface/affective-semantic features only, as could be expected theoretically. The top-ranked features being concreteness/conc, and aesthetic potential/ap suggests that aesthetic aspects also may play a role in similarity ratings; an interesting issue for future studies.

Most importantly, the type IV hybrid models all perform quite well on the training set, while mostly achieving cross-validated $R^2$s superior to those obtained by the other model types. The ‘winning’ model 28 (ERT) accounted for about 50% of the variance and
together with the type III model results, we take this as support for our working hypothesis that word similarity ratings are based not only on semantic relatedness but on other features as well. However, the feature importance values of the winning model also indicate that semantic relatedness and association –estimated by hybrid knowledge *and* corpus-based models like wn-lin or predict models like w2v– play the front role. The results of the hybrid models produced by the ERT regressor further indicate that the wn-lin measure also is of importance when many other features are added (it descriptively outranks even the w2vwiki300 feature in the winning model 28). The four models with –descriptively– the best performance in each class were models # 7, 12, 19, and 28, not astonishingly all being combinations of several models and having a maximum number of features. This may inspire more research trying to find the optimal number and combination of a large set of word features to predict similarity ratings.

The most striking result from Table 1, however, is the moderate cross-validated performance[6] of the hybrid models. This suggests that there must be other factors involved which drive ratings beyond the features considered here. An obvious source of unaccounted variance lies in the human raters and item variability. The rating data (SIMLEX999, https://www.cl.cam.ac.uk/~fh295/simlex.html) provide only mean values without standard deviations, thus leaving open the question which word pairs have the highest or lowest inter-rater agreement[7]. It would be interesting to repeat the present analyses on a subset of items for which the inter-rater variance is a minimum or to include this variance into the models.

[6] Lacking directly comparable data in the literature, it is difficult to evaluate the cross-validated performance of the present models. It should be noted though that the presumably ‘best’ current performance in accounting for human association ratings for word pairs by a computational model of association strength is in the range of $R^2$ = .37 - .47 (*without* cross-validation; Hofmann et al., 2018). When predicting original ratings about single word features (e.g., *valence* or *concreteness*) with ratings extrapolated via a random forest model trained on 25% of the full dataset (i.e., *with* cross-validation) the results varied between: $R^2$ = .10 - .6 (mean = .55, N = 45; Mandera et al., 2015; Table 1).

Another issue related to the cross-validated performance accounting for about half of the variance at best is feature selection[8]. We have discussed the general problem of (pre-)selecting the relevant or 'right' features used in predictive modeling of human rating data elsewhere (Jacobs, 2015, 2017; Jacobs & Kinder, 2017). Suffice it to say that state-of-the-art QNA tools like coh-metrix (Graesser et al., 2004), SEANCE (Crossley et al., 2016) or Natural Language Processing packages (NLTK etc.) offer hundreds of features and researchers interested in more than the few standard lexical features typically used in word recognition or reading research (e.g., word frequency, valence) are often at a loss –lacking appropriate theoretical model guidance– when having to decide which ones to chose for which reason. Clearly, more psychological and interdisciplinary research in this direction is of essence here. The 4x4 feature matrix developed in the emerging field of Neurocognitive Poetics (Jacobs, 2015, 2018b) offers a taxonomy of sublexical, lexical, interlexical and supralexical text features at the metric, phonological, morpho-syntactic and semantic levels for empirical studies of literature. Even if this may be useful for many future studies of reading, it does not specify which features are likely involved in a task consisting of judging the similarity of two words. While it is possible that the present selection and combination of features is not the optimal choice for this task, we know of no other studies that have looked into this issue and think that this is a promising start for further research challenging the present one. Similarly, the modest cross-validated performance of the type I and II models suggests that within the class of LSA and w2v models there also still room for improvement (cf. Faruqui et al., 2016).

The main point we'd like to make, however, is that the models used here cannot replace psychological process models of the similarity rating task, although they should motivate and help to develop them. Such process models (e.g., implemented as simulation models of the localist connectionist family including a semantic layer; Hofmann & Jacobs, 2014; Jacobs,

[7] The SIMLEX999 homepage indicates the average pairwise Spearman correlation between two human raters as being .67 without further specifications.

[8] In pilot studies, we used sklearn's *mutual_info_regression* –which in contrast to the F-test alternative can handle non-linear dependencies– for preselection of the present features among a set of >30 word features (e.g., *number of higher frequency neighbors* or *emotion potential* were excluded).

Graf, & Kinder, 2003) should specify, for instance, the salience of both the common and distinctive features of both words (cf. Gati & Tversky, 1984) or the density of features in a particular region of the n-dimensional space (cf. Krumhansl, 1978). At least some of the computational models considered here could be extended to implement both aspects. Alternatively, following our simulation model of lexicality ratings based on a measure of overall orthographic lexical activity (Jacobs et al., 2003), one could develop a localist connectionist model that –in line with the present working model– generates similarity ratings on the basis of three overall activities (orthographic, phonological and semantic lexica) using an extended orthographic lexicon of the Multiple Read-Out Model (MROM; Grainger & Jacobs, 1996), the phonological lexicon of the MROM-p (Jacobs et al., 1998) and the semantic lexicon of the AROM (Hofmann et al., 2011; Hofmann & Jacobs, 2014). However, further empirical data are required to constrain future modeling efforts, e.g. concerning which aspects human raters focus on, or experimentally manipulating some of the features considered here to obtain human data reflecting feature weights (cf. Keren & Baggen, 1981; Tversky, 1977). We view the present analyses as a start which is complex and successful enough to motivate further research –both theoretical and empirical– and to challenge the somewhat one-dimensional standard computational perspective on word similarity ratings which dominates the literature so far.

## Figures

**Figure 1. Results for best-fitting model (model 28 ERT; N = 222 randomly chosen test data set)**

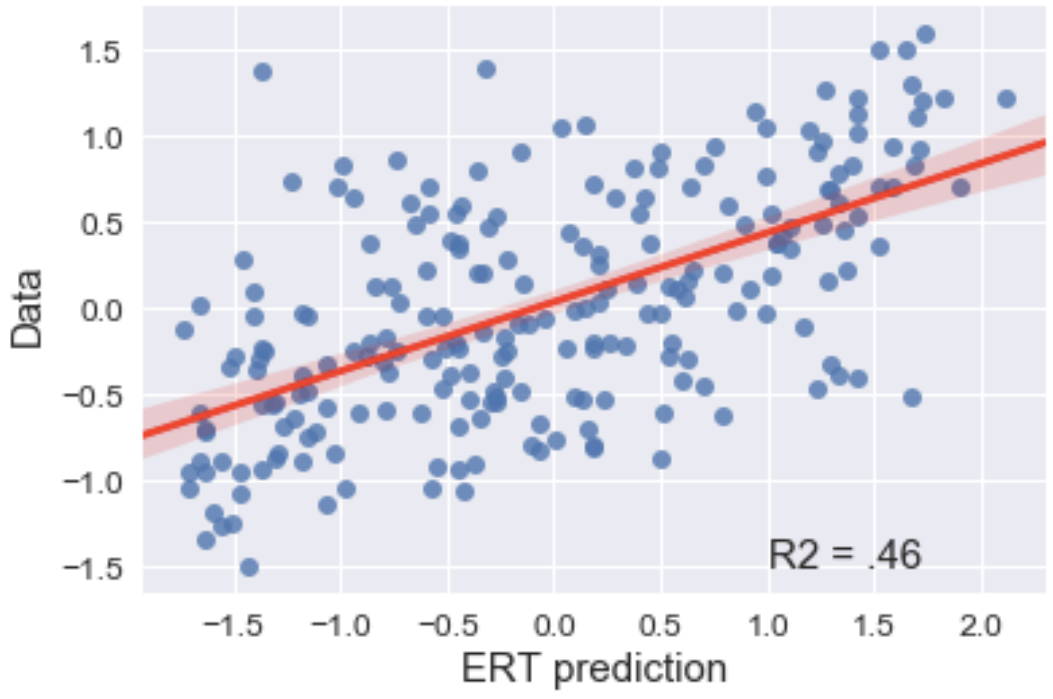